\documentclass[11pt,a4paper]{article}
\usepackage[hyperref]{acl2017}
\usepackage{times}
\usepackage{latexsym}
\usepackage{graphicx}
\usepackage{url}
\usepackage{stfloats}

\aclfinalcopy 

\title{Modeling the Complexity and Descriptive Adequacy \\
 of Construction Grammars
}

\author{Jonathan Dunn \\
  Illinois Institute of Technology \\
  Dept. of Computer Science \\
  {\tt jdunn8@iit.edu}}

\date{}

\begin{document}
\maketitle

\begin{abstract}
This paper uses the Minimum Description Length paradigm to model the complexity of CxGs (operationalized as the encoding size of a grammar) alongside their descriptive adequacy (operationalized as the encoding size of a corpus given a grammar). These two quantities are combined to measure the quality of potential CxGs against unannotated corpora, supporting discovery-device CxGs for English, Spanish, French, German, and Italian. The results show (i) that these grammars provide significant generalizations as measured using compression and (ii) that more complex CxGs with access to multiple levels of representation provide greater generalizations than single-representation CxGs.
\end{abstract}

\section{Complexity and Descriptive Adequacy}

Construction Grammars (CxGs; Goldberg, 2006; Langacker, 2008) operate at multiple levels of representation (lexical, syntactic, and semantic) making them potentially much more complex than purely syntactic grammars. This paper models both (i) the computational complexity of CxGs and (ii) their descriptive adequacy against unannotated corpora using Minimum Description Length (MDL). These two measures, complexity and descriptive adequacy, can be used together as an objective function for measuring the quality of CxGs: the optimum grammar balances higher descriptive adequacy against lower complexity. Once we can measure the quality of a particular grammar in reference to a corpus of observed language use, we can search until we find the optimum grammar for that corpus. This paper uses measures of complexity and descriptive adequacy to learn CxGs for English, Spanish, French, German, and Italian. 

The goal is not to examine the representational capacity of CxGs in general because CxG is a fundamentally usage-based paradigm (Hopper, 1987; Kay \& Fillmore, 1999; Bybee, 2006). This means that the general capacity of its grammars must be weighted by their actual content: how can we model the complexity of a specific CxG used to describe a specific language, where that language is represented by a specific observable corpus?

Previous computational work on CxG (Steels, 2004; Bryant, 2004; Chang, et al., 2012; Steels, 2012) has relied on introspection-based representations that require a linguist to determine the optimum constructions by intuition. From a linguistic perspective, these representations are neither replicable nor falsifiable and are unable to test hypotheses about the mechanisms of emergence that map from observed usage to learned generalizations. From a computational perspective, these representations are not scalable across domains and languages and are subject to all the constraints of knowledge-based systems. Other data-driven approaches (Wible \& Tsao, 2010; Forsberg, et al., 2014) generate potential constructions but do not evaluate the quality of competing CxGs as collections of constructions.

Section 2 discusses how CxGs are represented and Section 3 considers interactions between different levels of representation. Section 4 presents Minimum Description Length as a joint measure of complexity and descriptive adequacy suitable for measuring grammar quality while Section 5 operationalizes CxG encoding. Section 6 describes the search algorithm for optimizing grammar quality. Section 7 describes a multi-lingual experiment in measuring CxG complexity and descriptive adequacy and Appendix A discusses constructions learned from the corpus of English.

\begin{table*}[t]
  \centering
  \begin{tabular}{l}
(1a) [\textsc{slot 1} --- \textsc{slot 2} --- \textsc{slot 3} --- \textsc{slot 4}] \\
(1b) [\textsc{noun} --- ``gave" --- $(animate)$ ---  ``a hand"] \\
(1c) ``Bill gave Peter a hand." \\
(1d) [\textsc{noun} --- $(transfer)$ --- $(animate)$ --- \textsc{noun}] \\
(1e) ``Bill sent Peter a package." \\
  \end{tabular}
  \caption{Construction Notation and Examples}
  \label{tab:1}
\end{table*}

\section{Representing CxGs}

 This section introduces the symbolic notation used to represent CxGs and describes how these representations are implemented. The algorithm recognizes three distinct types of representation as atomic units in its descriptions: Lexical representation consists of tokenized word-forms (in lowercase). Syntactic representation consists of part-of-speech categories (defined using the Universal POS tagset, Petrov, et al., 2012, and computed using RDRPosTagger, Nguyen, et al., 2016). Semantic representation consists of clusters of distributionally similar words that represent semantic domains and are computed using GenSim's implementation of word2vec (Rehurek \& Sojka, 2010). The embedding model is trained using 1 billion words from web-crawled corpora for each language (from the WAC corpora: Baroni, et al., 2009; and Aranea corpora: Benko, 2014) using skip-grams with 500 dimensions. These embeddings are segmented into categorical domains using k-means clustering ($k = 100$). The idea behind these three types of representation is that a particular slot in a construction can be defined or constrained at the lexical, syntactic, or semantic level. These representations thus form the basic alphabet of the algorithm.

Constructions are sequences containing a certain number of slots, as in (1a) with four individual slots. Each construction is surrounded by brackets and each slot within a construction is separated by a dash (``---"). Each slot in a construction is represented or defined by constraints that govern which units can occupy that slot. Lexical constraints are indicated using single quotes (e.g.,``gave" in 1b). Syntactic constraints are indicated using part-of-speech tags in uppercase (e.g., \textsc{noun} in 1b). Semantic constraints are indicated within parentheses with the identifier for the semantic domain (e.g., $animate$ in 1b). Thus, the construction in (1b) describes the utterance in (1c) but not the utterance in (1e); the construction in (1d) describes the utterances in both (1c) and (1e).

This provides a good example of the complexity problem: CxGs potentially have multiple overlapping representations for any given sentence. The sentence in (1e), for example, can be represented by the construction in (1d), in which slots are defined by both syntactic constraints (i.e., \textsc{noun}) and semantic constraints (i.e., $animate$). CxGs can distinguish between (1e) and its more idiomatic counterpart (1c) using representations such as (1d) and (1b). The question, however, is how many of these item-specific or idiomatic representations are needed in the grammar: each item-specific construction increases grammar complexity.

In this paper, the term {\em construction} refers to the grammatical description (e.g., 1b) and the term {\em construct} refers to a member of the set of utterances which that construction represents (e.g.,1c). For a given grammar, the set of constructions is closed but the set of constructs is open. A construct or utterance can be represented by multiple constructions: representations like (1b) that are more item-specific alongside representations like (1d) that are more schematic. This leads to relationships between constructions: an inheritance hierarchy in which (1b) is a child of (1d). The current implementation has three limitations in respect to the ideal CxG: First, constraints are limited to a single type of representation per slot. For example, if a slot is constrained to the semantic domain $animate$, any syntactic category could be used to fill that slot. Second, although constituents are able to fill construction slots (i.e., ``a hand" can occupy a single slot as a single \textsc{noun}), larger constructions such as (1d) cannot fill slots in other constructions. Third, no relations are learned between constructions in the grammar (i.e., the inheritance hierarchy is not modeled). 

In computational terms, each construction is an array of slots. Each slot is defined as a tuple that contains two pointers: first, a pointer to the alphabet constraining that slot (i.e., lexical or syntactic units) and, second, a pointer to a particular unit within that alphabet (i.e., ``a hand" or \textsc{noun}). Constituents are allowed to fill slots. This is accomplished using a context-free phrase structure grammar containing rules such as

 $$\textsc{determiner --- noun} \rightarrow \textsc{noun}$$

 that is learned during a syntax-only iteration described in the next section. The syntactic alphabet, then, also supports pointers to complex sequences through this CFG: the construction points to a \textsc{noun} and the CFG allows larger constituents to be labeled as a single \textsc{noun}. The current implementation produces a context-free CxG.

\section{Finding CxGs}

In the experiments that follow, each language is represented by a large web-crawled corpus in that language. Its grammar is learned by searching across potential grammars, each of which is evaluated against the corpus until the optimum grammar is found (using a measure defined in Section 4). The search for the optimum grammar is conducting using a tabu search (Glover 1989, 1990a) with multi-unit association measures (Dunn, 2017) used to sample potential constructions. The main focus of this paper is on defining the objective function: how can we know that one grammar is better than another without evaluating them against gold-standard annotations?

Three levels of CxGs are learned: The first pass operates on only lexical representations, $CxG_{LEX}$. This identifies purely lexical constructions: sequences of lexical items that have been fused together so that their internal structure can be ignored. For example, ``could be" and ``will be" are identified as single units when the algorithm is applied to English. Later passes view these lexical constructions as a single lexical item with a single syntactic type. 

The second pass operates on only syntactic representations, $CxG_{SYN}$. Syntactic constructions are later used as phrase structure rules. For example, when applied to English the sequence 

$$[\textsc{verb} - \textsc{noun}]$$

\noindent is identified as a purely syntactic construction. In later passes, these sequences are converted into constituents that can be treated as a single unit. The third pass operates on all levels of representation, $CxG_{FULL}$. The grammar accumulates structure across these iterations in the sense that constructions output from a previous pass become atomic units in the current pass. This set-up allows us to examine complexity and descriptive adequacy across CxGs with access to different levels of representation: do we actually benefit from more complex multi-level grammars?

\section{Measuring Grammar Quality}

This approach depends on the central insight of MDL (Rissanen, 1978, 1986; Gr{\"u}nwald \& Rissanen, 2007): a grammar is a method for encoding observed linguistic utterances and the learner is searching for the smallest adequate encoding method. Explanation here is a matter of prediction: can the grammar produce the utterances observed in held-out test-sets? The optimum grammar balances model complexity (the number and type of constructions in the grammar) and the amount of compression achieved when the model is used to encode a test corpus (c.f., Goldsmith, 2001; 2006). The complexity of the grammar is balanced against its descriptive adequacy on a held-out corpus. This is formalized in MDL as 

$$MDL = \min\limits_{G} \{⁡{L_1(G)+L_2 (D \mid G)}\}$$

This defines the optimum grammar as the one which minimizes the model complexity, represented by the encoding size of the grammar, plus the size of the dataset encoded by means of the grammar. Encoding size in MDL (here based on the natural log) is further defined as

$$L_C (X^n )=-〖log〗_e⁡P(X^n )$$

Methods for calculating the encoding size of CxGs are discussed below in Section 5.  An additional term, $L_3(G)$, is sometimes used (Gr{\"u}nwald \& Rissanen, 2007: 409) to control for the size of the encoding required for the universal code used to determine the size of $G$. This term is often not included in the MDL metric (it is not necessary when evaluating models against one another). It will be necessary here, however, when measuring grammar quality against the baseline of an unencoded test set. We are using the MDL principle as a metric for model selection. One aspect of model selection is confidence: to what degree is $G_A$ better than $G_B$? This is given by

$$|MDL(G^A )-MDL(G^B )|$$

Higher values indicate a more significant difference between $G_A$ and $G_B$ (c.f., Gr{\"u}nwald \& Rissanen, 2007: 411). This measure of confidence will be useful for evaluating the quality of grammars against the baseline of an unencoded dataset. We can refine this measure of confidence further

$$1- \frac{MDL(G^A )}{MDL(U)}$$

\begin{figure*}
\centering
\includegraphics{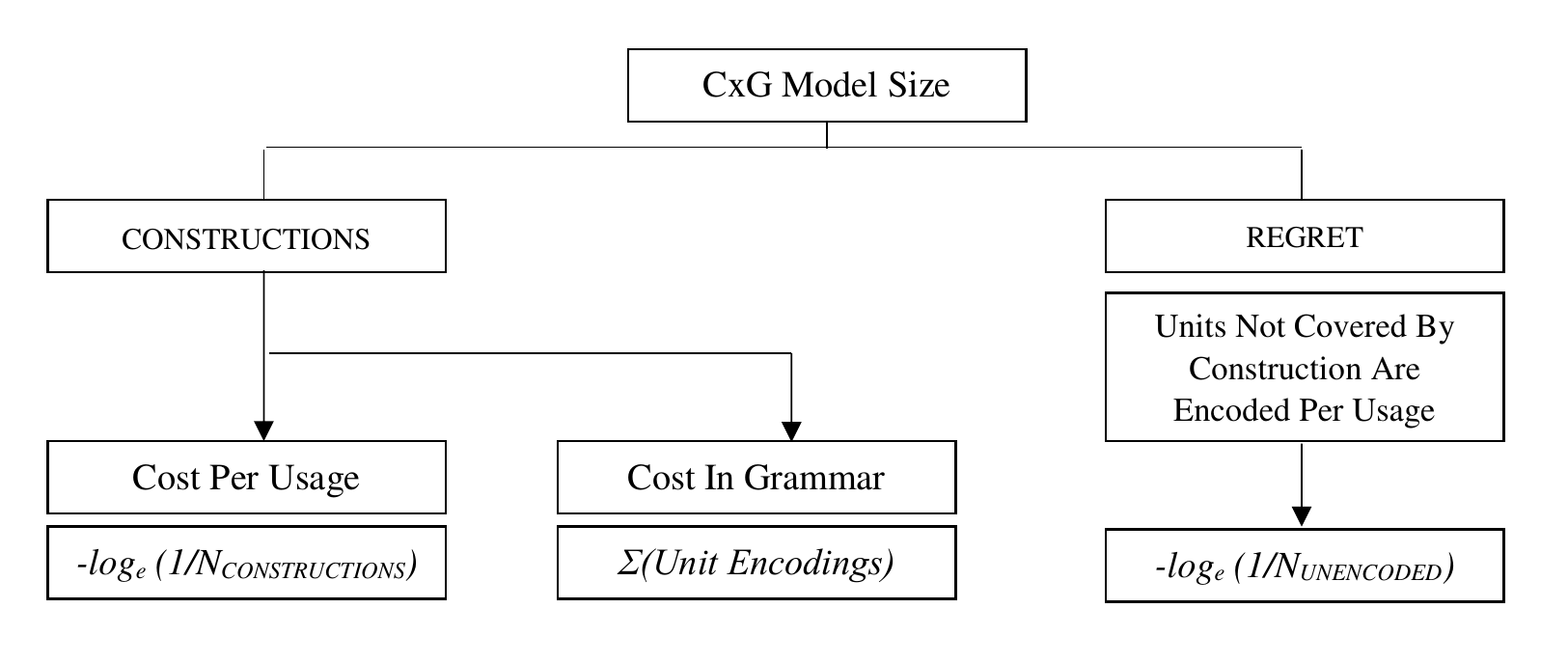}
\caption{Encoding Model}
\end{figure*}

This is the relative degree of compression adjusted so that values close to 1 represent higher compression ($U$ represents the size of the data without the grammar). Thus, if the MDL of $G_A$ is 256 and the unencoded MDL, $U$, is 927, this gives a compression of 0.7239 over the unencoded baseline as a measure of grammar quality. Negative values indicate that the grammar makes the MDL metric worse, an unlikely but possible occurrence. This ratio measure is important because, without it, the MDL metric and the significance of the metric are both dependent on the encoding size of a specific test set.

We are searching for the grammar with the lowest MDL metric on a held-out test set, but we also need to measure the amount of variation across restarts. This provides a measure of stability: a restart is a search technique that restarts the search for the optimum grammar from scratch on a different portion of the corpus in order to determine if similar grammars are discovered. Let $G_k$ be the optimum grammar across restarts and $G_{i\ldots n}$ be the set of all output grammars across restarts regardless of whether they are optimal. The agreement between the two grammars is

$$A_i^k = \frac{(k\cap i)}{(k\cup i)}$$

The significance of the difference between the encoding quality of $k$ and $i$ relative to the encoding quality of the optimum grammar is

$$M_i^k= 1- \frac{|MDL(k)-MDL(i)|}{MDL(k)}$$

This is adjusted to make large differences closer to 0 and small differences closer to 1. The stability measure, $STA(k)$, is 

$$\frac{\sum_{i=0}^{n}▒〖A_i^k (M_i^k )}{n}$$

This is the mean agreement between the current grammar and the optimum grammar for all restarts $n$ with each weighted so that more significant differences in the MDL metric lower the agreement. This is a joint measure of stability in grammar content and grammar quality, with higher scores (toward 1) indicating stable grammars and lower scores (toward 0) indicating unstable grammars.

This section has used the Minimum Description Length paradigm to develop measures of grammar complexity (i.e, $L_1$) and descriptive adequacy (i.e., $L_2$) that do not rely on gold-standard annotations. This is important for two reasons: First, we do not necessarily have gold-standard annotations for every language and language variety we are interested in (i.e., CxGs are also subject to variation). Second, simply relying on gold-standard annotations ignores the question we are most interested in: how do we know empirically that one grammar is better than another?

\section{Measuring the Encoding Size of CxGs}

 The MDL paradigm depends on the concept of encoding size to measure complexity and descriptive adequacy: how do we calculate this for CxGs? The MDL metric contains three terms: $L_1$, or the encoding size of the grammar; $L_2$, or the encoding size of the corpus given the grammar; and $L_3$, or the encoding size of the universal code necessary for encoding $L_1$. Additionally, we need to determine the uncompressed encoding size of the corpus to serve as a baseline for measuring the overall rate of compression of competing grammars.

The basic encoding model, shown in Figure 1, has two top level categories composing its alphabet: $Constructions$ (representations within the current CxG), and $Regret$ (units not described by known constructions). Each of these top-level categories is assigned the same probability, 0.5, and thus, because encoding size is equivalent to  $-〖log〗_e⁡P(X^n )$, each comes with an initial encoding size of 0.693 nats (where a {\em nat} is a {\em bit} based on the natural logarithm). 

The reason for separating these top-level categories is that each has a different number of units, each of which is again assigned equal probability. For example, if there are 1,000 constructions in the grammar, then each usage of a construction costs 0.693 nats (for indicating a construction) and 6.907 nats (for pointing to a specific construction). Rather than assume that each construction in a given CxG is equally probable, an alternate approach is to assign probabilities to individual constructions and use these to determine the cost of encoding constructions on an individual basis. This problem is left for future work. Here, constructions are distinguished from one another only using (i) their relative complexity and (ii) the productivity of the particular grammar they belong to.

The $Regret$ category holds units in the corpus that are not described by a construction in the current grammar. Each occurrence of a non-construction unit is encoded on-the-fly: as the number of undescribed units increases, the cost in nats of encoding each occurrence also increases. For example, if there are 1,000 undescribed units the cost per unit is 0.693 nats plus 6.907 nats; if there are 10,000 undescribed units the cost per unit is 0.693 nats plus 9.210 nats. This cost is specific to a given dataset, not to a given model, because the cost per undescribed unit depends on the total number of undescribed units. It is important to note that each instance of a unit not described by the grammar is stored in the $Regret$ category independently: this is a measure of model error.

If the encoded dataset were transmitted, the model itself would need to be encoded and transmitted at the same time in order to decode the dataset; this is the information-theory rationale behind $L_1$, the encoding size of the grammar. In linguistic terms, grammars with larger encoding sizes are more complex. The $Regret$ category has already been encoded with unique pointers for each undescribed unit; thus, it does not incur an additional model cost. The cost of encoding the model, then, consists entirely of the cost of encoding each construction it contains: the sum of all unit-encoding costs for each slot-filler representation in the construction,

$$\sum_i^{N_{SLOTS}}▒〖-log_e(\frac{1}{N_{R_{i}}} ) 〗+ -log_e(\frac{1}{T_R})$$

$N_{SLOTS}$  here is the number of slots in the construction being encoded, $N_{(R_i )}$ is the number of units available for a given representation type, and $T_R$ is the number of representation types total for the current grammar. This is the total cost of encoding both (i) which representation type (alphabet) fills the slot and (ii) which unit of that alphabet fills the slot. 

The full CxG has three representation types so that, for this grammar type, the encoding size for each slot is 1.098 nats (the cost of encoding a three-way distinction) plus $–log_e(1/N)$ where $N$ is the total vocabulary of that unit type. Thus, if there are 20,000 lexical items in the vocabulary, the cost of encoding a construction with three lexically-filled slots is 11.001 nats per slot or 33.003 nats total. This is a one-time encoding cost: each occurrence of a construction is a pointer that incurs the encoding cost described above.

The $Regret$ category more properly belongs as an added term in $L_2$, the size of the dataset as encoded by the grammar. However, in this case it clarifies the discussion of grammar complexity to show the impact that each unencoded unit has on the MDL metric as a whole. Note that the complexity cost includes $L_3$ or the cost of encoding the encoding size of the grammar. In other words, in order for each construction to be encoded we also have to encode the lexicon of lexical, syntactic, and semantic units used in construction descriptions. This is included as part of the cost of each construction in the grammar.

\begin{figure*}
\centering
\includegraphics{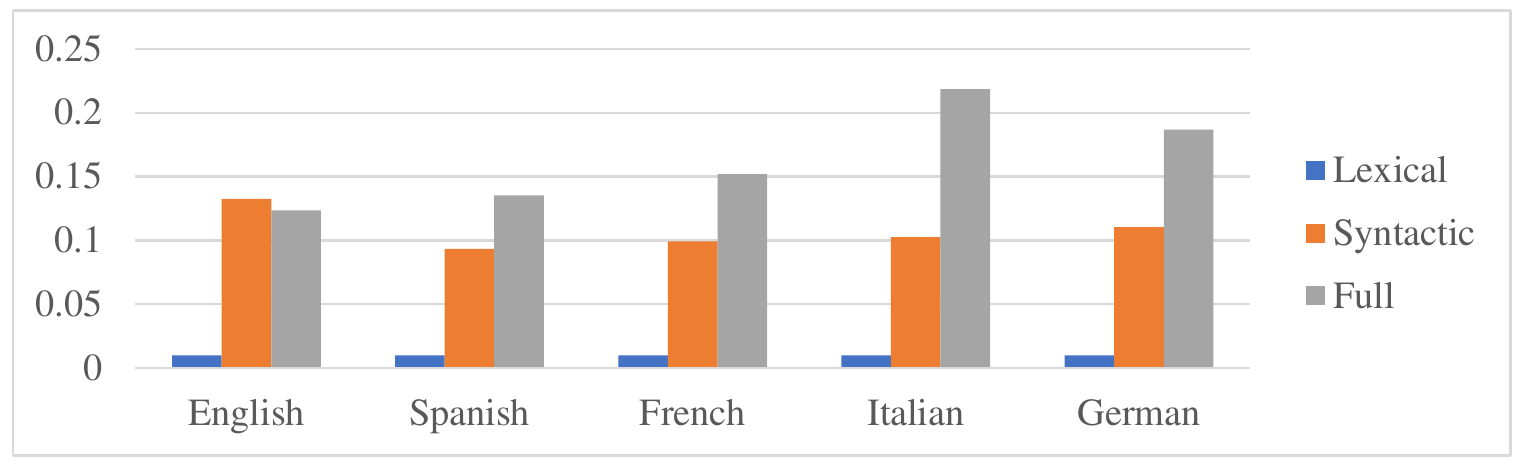}
\caption{Compression Rates Across Grammar Types}
\end{figure*}

\subsection{Maintaining Lossless Encoding}

The data consists of atomic units from three types of representation. In order to maintain the grammar as a lossless encoding of the corpus, we define the task for $CxG_{FULL}$ as encoding one of these representations for each unit. Importantly, this means that each unit needs to be represented by only one type of representation in the decoded version of the dataset; part of the learning task for $CxG_{FULL}$ is to choose the optimum type of representation for each slot. What lossless encoding means, in practice, depends on the type of CxG. For $CxG_{LEX}$ lossless encoding means to return the same word-forms but for $CxG_{SYN}$, it means to return the original sequence of parts-of-speech.

It is important that $CxG_{SYN}$ is evaluated while not in Chomsky normal form in order to correctly encode the complexity of the grammar.  Consider the individual phrase structure rules in (2a) through (2c) which map from a particular sequence of part-of-speech tags to a single constituent type. For $CxG_{SYN}$ internally each sequence is a construction (i.e., phrase structure rules are not typed). In the same way, for $CxG_{LEX}$ internally, each sequence of word-forms is not typed (i.e., assigned to a part-of-speech category). Constructions from each of these passes need to be typed before filling slots in later passes. The part-of-speech tagger is used to assign lexical constructions to a single part-of-speech. An additional algorithm (outside the scope of this paper but available in the external resources) converts $CxG_{SYN}$ sequences into phrase structure rules to support the CFG that allows longer sequences to fill individual slots.

~

(2a) \textsc{det} --- \textsc{noun} $\rightarrow$ \textsc{noun}

(2b) \textsc{noun} $\rightarrow$ \textsc{noun}

(2c) \textsc{noun} --- \textsc{noun} $\rightarrow$ \textsc{noun}

~

The point is that, while the representations in (2a) through (2c) do not provide a lossless encoding of the observed utterances, the MDL metric is not applied to these representations but to their untyped forms (e.g., [\textsc{det} --- \textsc{noun}]). The CxG encoding system consists of $Atomic Units$ located within $Constructions$. As the level of abstraction increases (i.e., as we go through multiple iterations), members of the $Construction$ repository for the current pass become members of the $Atomic Units$ repository for the next pass. Thus, lexical constructions are considered part of $Constructions$ in $CxG_{LEX}$ but part of $Atomic Units$ in $CxG_{SYN}$. The effect of this is to maintain lossless encoding at each level of abstraction while incorporating previously learned representations into the next level of abstraction.

This means that grammar complexity is not directly comparable across iterations because each iteration is encoding a different level of abstraction. For example, the task for $CxG_{SYN}$ is to provide a lossless encoding of sequences of syntactic units (out of an inventory of 14 unit types). A relatively small number of syntactic sequences will be able to form phrase structure rules that, taken together, provide a high rate of compression. A full CxG, however, must do much more than predict sequences of syntactic units because it also incorporates lexical and semantic representations. On the other hand, though, the full MDL metric is comparable across iterations because it balances complexity and descriptive adequacy: does the more complex $CxG_{FULL}$ provide enough descriptive adequacy to justify incorporating multiple types of representation?

\begin{figure*}
\centering
\includegraphics{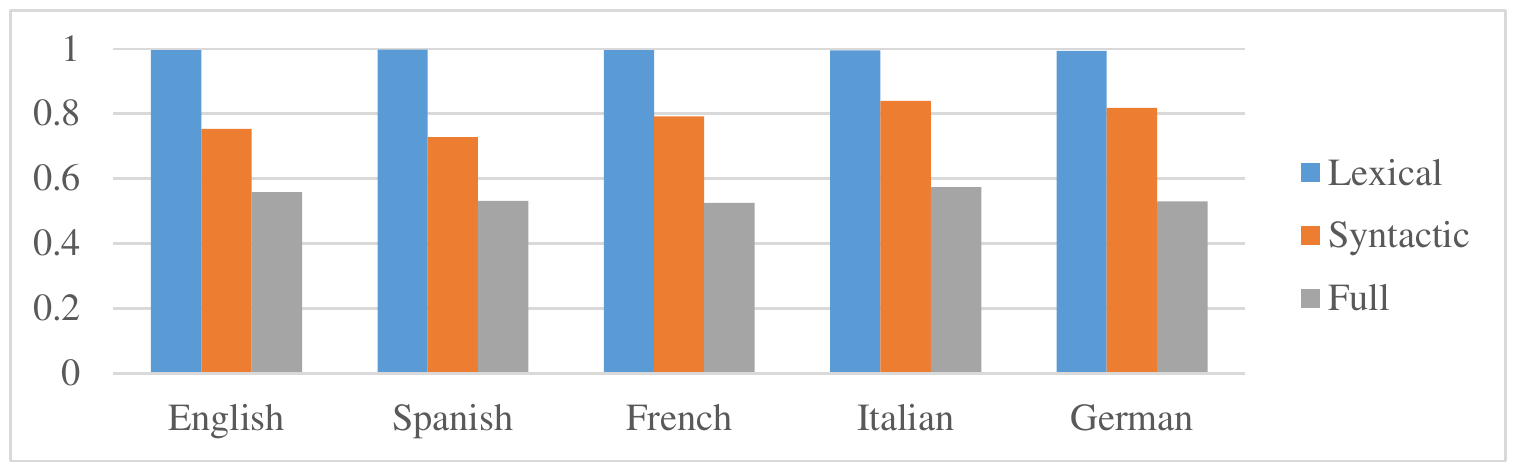}
\caption{Stability Over Folds Across Grammar Types}
\end{figure*}

\section{Searching Over Potential CxGs}

The search algorithm has three components: (i) randomly initializing the starting state, or what set of constructions belongs in the initial grammar; (ii) an indirect tabu search (Glover, 1989, 1990a) to move toward the optimum grammar by updating construction sampling parameters; and (iii) a direct tabu search across constructions to determine if small changes in the inventory of constructions improves the quality of the current CxG.

The first tabu search takes a randomly initialized starting state and searches for improved grammars by exploring different sampling parameters. These parameters take a number of association measures (c.f., Dunn, 2017) and use them to determine which constructions belong in the grammar. The essential idea of tabu search is (i) to define the set of possible moves from the current grammar state to a new grammar state and (ii) to combine tabu restrictions and aspiration criteria to move the search toward promising areas of the overall search space that are not directly reachable from the current state. We divide the parameter space into $n$ discrete values for each of the 30 direction-specific association measures, with the maximum and minimum values defined empirically. This provides a finite set of possible moves from any given state.

For each turn, the algorithm generates a set of possible moves and, after evaluating each, takes the best available move even if it reduces the overall grammar quality. \textit{Best} is defined using the MDL-metric: the best move is that which provides the smallest MDL-metric of all possible moves. \textit{Available} is defined as a move that is either (i) not on the tabu list or (ii) satisfies an aspiration criteria that overrules the tabu list. The tabu list is a short-term memory item that contains the last $n$ moves, each represented using the association measures that have been changed. For practical reasons, $n$ is set at 7 (c.f., Glover, 1990b); this means that for any given turn the best move cannot involve a sampling parameter that has been changed over the last 7 turns. This prevents the algorithm from cycling between local optima in an endless loop. The aspiration criteria used is that the grammar produced by a move is not only the best available grammar but also the best observed grammar: a new global minimization of the MDL metric. Thus, the tabu against altering a recently changed sampling parameter can be overruled if that change creates a new best grammar. The use of such an aspiration criteria makes intuitive sense: the tabu search is designed to prevent cycling between previously visited states, but a grammar which reaches a new global minimum has not previously been visited.

Three types of moves are available at each turn: First, a parameter can be removed from the current sampler (i.e., \textsc{off}); this allows the tabu search to eliminate sampling parameters that reduce grammar quality. Second, a parameter can be grouped with $n$ randomly chosen changes to other parameters (i.e., \textsc{and}); this allows the tabu search to explore states similar to the current grammar. Third, a parameter can be allowed to overrule all other parameters (i.e., \textsc{or}); this allows the tabu search to move toward better but more distant states. 

Potential moves for each turn are generated as follows: for each association measure, one move is added with that measure removed from the sampler (\textsc{off}); two \textsc{or} moves above and two below the feature's current threshold, serving as escape hatches; and 25 \textsc{and} moves that include the current feature and $1...k$ other features ($k=5$). The stopping criteria is that a new best grammar has not been observed for 14 turns, twice the size of the tabu list. This stopping criteria is an intermediate memory item that monitors the general direction of the search. The intuition is that, if a new optimum grammar has not been reached within two complete cycles of the short-term tabu list, such a grammar is unlikely to exist. It is important to keep in mind that each turn evaluates a wide range of possible moves. This means that a large number of potential grammars are evaluated in determining each move. Given the size of the space reachable from any given state and the number of states visited during the tabu search, it is unlikely at this point that a significantly better grammar exists.

\section{Results and Discussion}

The evaluation uses web-crawled corpora (from the WaC and Aranea projects) for English, Spanish, French, German, and Italian. The same data segmentation shown in Table 3 is used for each language. Each grammar type is evaluated using cross-validation with two folds; the training-testing split is randomly assigned. The search stage uses two restarts, each with a unique segment of the training data. This means that the learning algorithm makes four passes per iteration (two folds with two restarts) over which we can measure stability.

\begin{table}[h]
  \centering
  \begin{tabular}{ll}
Used For & \# Sentences \\
\hline
Candidates and Association & 1 million \\
Test Sets for Restarts ($Lex, Syn$) & 100k \\
Test Sets for Restarts ($Full$) & 20k \\
Calculating Evaluation Metric & 200k \\
\end{tabular}
  \caption{Data Segmentation Per Fold}
  \label{tab:1}
\end{table}

The first measure of grammar quality, in Figure 2, is the compression achieved over the unencoded dataset on held-out testing data. Values close to 1 represent a large amount of compression while values close to 0 represent very little compression. We see across languages that lexical constructions (i.e., ``because of") do not provide much compression. In part, this is because few such constructions are selected: an average of 22 per language. Purely syntactic constructions, however, do provide compression (with an average of 120 identified per language). For all languages except English $CxG_{FULL}$ has the highest rate of compression, with each grammar containing between 4k and 5k constructions.

This measure shows us three things: First, there is a balance between complexity and descriptive adequacy that is forced by MDL. In other words, the descriptive power of the purely lexical constructions are only able to justify 22 constructions as opposed to 4k - 5k for $CxG_{FULL}$. CxGs with multiple types of representations are allowed to produce more complex grammars because they produce better descriptions of the corpus. Second, the learned grammars provide meaningful generalizations. In other words, this compression metric shows that not only does the algorithm find the optimum grammar with respect to competing grammars but that it also finds grammars that offer above-the-baseline compression. Full compression is, of course, impossible and these results provide a benchmark for future work. Third, these results show that the addition of semantic representations provide improved descriptive adequacy. A representative sample of the output of $CxG_{FULL}$ for English is shown in Appendix A.

How consistent are the grammars learned across different sub-sets of the corpora? This is shown in Figure 3 using the stability metric introduced in Section 4 over the grammars produced from different sub-sets of the corpora. We see that more complicated grammars are less stable. Thus, $CxG_{LEX}$ has low compression but almost perfect stability because the same small number of lexical constructions are consistently identified. $CxG_{FULL}$, on the other hand, has a much larger number of constructions that provide much higher compression; but the inventory of these constructions is subject to more variation. 

\begin{table*}[!b]
  \begin{tabular}{ll}
\textbf{Appendix A: Representative Examples} \\
\hline

{[\textsc{adverb} --- ``about"]} & {\em Modified Adverbs}\\ 
\hspace{5mm} ``at about"  & This simple construction modifies adverbs  \\
\hspace{5mm} ``how about" & to include information about vagueness. \\
\hspace{5mm} ``only about" &  \\
\hspace{5mm} ``on about"  \\
\hline
 {[``provide" --- $25$ --- $25$]} & {\em Verb-Specific Direct Object} \\ 
\hspace{5mm} ``provide added value"  & This  verb-specific construction constrains\\
\hspace{5mm} ``provide an opportunity" &  the object of ``provide" to members of an \\
\hspace{5mm} ``provide general advice" & unlabeled semantic domain. \\
\hspace{5mm} ``provide information about" \\
\hline
{[$25$ --- ``to" --- $14$]} & {\em Complex Verb Phrase} \\
\hspace{5mm} ``designed to ensure"  & This construction represents a complex event  \\
\hspace{5mm} ``want to improve" & phrase that contains both a main verb, ``want," \\
\hspace{5mm} ``made to ensure" & as well as an infinitive verb, ``improve." \\
\hspace{5mm} ``able to understand" \\
\hline
{[\textsc{verb} --- ``to" --- $25$ --- \textsc{adverb}]} & {\em Evaluative Verb Phrase} \\
\hspace{5mm} ``need to consider how" & This construction describes a basic verb phrase \\
\hspace{5mm} ``wish to consider how" & embedded within an evaluative verb describing \\
\hspace{5mm} ``want to be here" & how the speaker perceives the event. \\
\hspace{5mm} ``like to find out" \\
\hline
{[\textsc{determiner} --- \textsc{noun} --- \textsc{adposition} --- $14$]} & {\em Complex Noun Phrase} \\
\hspace{5mm} ``some experience in research" & This construction encodes a noun phrase that \\
\hspace{5mm} ``a need for research" & contains a modifying prepositional phrase. \\
\hspace{5mm} ``the process of planning" \\
\hspace{5mm} ``a number of activities" \\
\hline
{[\textsc{sub-conj.} --- $25$ --- \textsc{adjective} --- \textsc{noun}]} & {\em Subordinated Noun Phrase} \\
\hspace{5mm} ``whether small independent companies" & This construction provides sub-ordinated \\
\hspace{5mm} ``that the international community" & noun phrases that attach to main clause verbs \\
\hspace{5mm} ``because the current version" & and then act as the subject for additional \\
\hspace{5mm} ``while the other party" & modifying material that remains unspecified.\\
\hline
{[\textsc{pron.} --- \textsc{aux.} --- \textsc{verb} --- \textsc{particle} --- $25$]} & {\em Partial Main Clause} \\
\hspace{5mm} ``you should continue to receive" & This construction represents the largest \\
\hspace{5mm} ``i was told to make" & representations that are identified by the \\
\hspace{5mm} ``they were going to have" & algorithm; it specifies most of a main clause \\
\hspace{5mm} ``this was going to be" & with a pronominal subject. \\
  \end{tabular}
\end{table*}

Lack of stability here is not necessarily caused by error: grammars are subject to variation. Some amount of this variation results from errors: tagging errors, parsing errors, and learning errors in which the search algorithm does not converge on the best grammar. Some amount of this variation, however, comes from differences in usage across different portions of the corpus: these large corpora contain many varieties, dialects, domains, and speakers, each introducing variant constructions. To what degree do these variations represent error and to what degree do they represent actual grammatical differences across the corpora? That is a question for future work because it requires testing grammars over data explicitly drawn from different varieties of a language.

This paper has shown that the MDL paradigm can be used to jointly model the complexity and descriptive adequacy of CxGs against unannotated corpora. This is important because methods that rely on gold-standard annotations to evaluate grammar quality ultimately depend on the introspections behind those annotations. How valid are these CxGs using external measures? One application-specific evaluation of a learned grammar is its ability to model dialectal variations. Separate work using these learned CxGs for dialectometry (Dunn, {\em Forthcoming}) shows that these grammars are able to model regional varieties with a high degree of accuracy. 

~

\footnotesize{\textbf{Resources.} Code and models for this work are available at \href{http://www.jdunn.name}{jdunn.name} and \href{https://github.com/jonathandunn/c2xg}{github.com/jonathandunn/c2xg}}

~

\footnotesize{\textbf{Acknowledgements.} This research was supported in part by an appointment to the Visiting Scientist Fellowship at the National Geospatial-Intelligence Agency administered by the Oak Ridge Institute for Science and Education through an interagency agreement between the U.S. Department of Energy and NGA. The views expressed in this presentation are the author's and do not imply endorsement by the DoD or the NGA.}

\clearpage

\bibliography{acl2017}
\bibliographystyle{acl_natbib}

\end{document}